\documentclass{article}

    \usepackage[final]{mlwg_2019}

\usepackage[utf8]{inputenc} % allow utf-8 input
\usepackage[T1]{fontenc}    % use 8-bit T1 fonts
\usepackage{hyperref}       % hyperlinks
\usepackage{url}            % simple URL typesetting
\usepackage{booktabs}       % professional-quality tables
\usepackage{amsfonts}       % blackboard math symbols
\usepackage{nicefrac}       % compact symbols for 1/2, etc.
\usepackage{microtype}      % microtypography
\usepackage{graphicx}       % include illustrations

\usepackage{mathtools}
\usepackage{amsmath,amssymb}
\usepackage{amsthm}
\usepackage{xcolor}

\newtheorem{theorem*}{Theorem}[section]
\theoremstyle{definition}

\newtheorem{definition*}{Definition}[section]

\newtheorem{remark*}{Remark}[section]

\newtheorem{example*}{Example}[section]

\title{Safe Predictors for Enforcing\\ Input-Output Specifications}

\author{%
  Stephen Mell,\hspace{.1cm} Olivia Brown,\hspace{.05cm} Justin Goodwin and Sung-Hyun Son\\
  MIT Lincoln Laboratory \\  
  sm1@seas.upenn.edu, \{olivia.brown, jgoodwin, sson\}@ll.mit.edu \\
}

\begin{document}

\maketitle

\begin{abstract}
  We present an approach for designing correct-by-construction neural networks (and other machine learning models) that are guaranteed to be consistent with a collection of input-output specifications before, during, and after algorithm training. Our method involves designing a constrained predictor for each set of compatible constraints, and combining them safely via a convex combination of their predictions. We demonstrate our approach on synthetic datasets and an aircraft collision avoidance problem.  
\end{abstract}

\section{Introduction}
  
  The increasing use of machine learning models, such as neural networks, in safety-critical applications (e.g., autonomous vehicles, aircraft collision avoidance) motivates an urgent need to develop safety and robustness guarantees. Such models may be required to satisfy certain input-output specifications to ensure the algorithms adhere to the laws of physics, can be executed safely, and can be encoded with any \emph{a-priori} domain knowledge.  In addition, these models should exhibit adversarial robustness, i.e., ensure outputs do not change drastically within small regions of an input -- a property that neural networks often violate (\cite{szegedy2013intriguing}).  %Ensuring these models exhibit such properties, however, has currently proven to be a technical challenge. 
  
  Recent work has demonstrated an ability to formally verify input-output specifications as well as adversarial robustness properties of neural networks.  For example, the Satisfiability Modulo Theory (SMT) solver Reluplex (\cite{katz2017reluplex}) was used to verify properties of networks under development for use in the Next-Generation Aircraft Collision Avoidance System for Unmanned aircraft (ACAS Xu), and later used to verify adversarial robustness properties in~\cite{katz2017towards}. While Reluplex and other similar approaches are successful at \emph{identifying} whether a network satisfies a given specification, these works provide no way to \emph{guarantee} that the network meets those specifications.  Thus, additional methods are still needed to modify networks if and when they are found to lack a desired property.
  
  Techniques for designing networks with certified adversarial robustness are beginning to emerge (\cite{Gowal2018effectiveness, cohen2019certified}), but enforcing more general safety properties in neural networks remains largely unexplored.  In~\cite{lin2019art}, the authors propose a technique for achieving provably correct neural networks through abstraction-refinement optimization and demonstrate their approach on the ACAS-Xu dataset. Their network, however, is not guaranteed to meet the desired specifications until after it has undergone training.  We seek to design networks for which input-output constraints are enforced even before the network has been trained to enable use in online learning scenarios, where a system may be required to guarantee a set of safety constraints are never violated during the entirety of its operation. 
  
  This paper proposes an approach for designing a \emph{safe predictor} (a neural network or any other machine learning model) that obeys a set of constraints on the input-output relationships, assuming the constrained output regions can be formulated to be convex.  Even before training begins, and at each subsequent iteration of training, our correct-by-construction safe predictor is guaranteed to meet the desired constraints. We describe our approach in detail in Section~\ref{Sec:Method}, and demonstrate its use for the aircraft collision avoidance problem from~\cite{julian2019verifying} in Section~\ref{Sec:Experiment}.  Results on synthetic datasets are shown in Appendix~\ref{app:synthetic}.

\section{Method} \label{Sec:Method}

Given two normed vector spaces, an input space $X$ and output space $Y$, and a collection of $c$ different pairs of input-output constraints, $(A_i,B_i)$, where $A_i \subset X$ and $B_i$ is a convex subset of $Y$ for each constraint $i$, the goal is to design a {\it safe predictor}, $F: X \rightarrow Y$, that guarantees $x \in A_i \Rightarrow F(x) \in B_i$.

Let $\vec{b}$ be a bit-string of length $c$. We define $O_{\vec{b}}$ to be the set of points $x$ such that, for all $i$,  $b_i = 1$ implies $x \in A_i$ and $b_i = 0$ implies $x \notin A_i$. $O_{\vec{b}}$ thus represents the overlap regions for each combination of the input constraints. For example, $O_{101}$ is the set of points in $A_1$ and $A_3$ but not in $A_2$, and $O_{0\cdots0}$ is the set where no input constraints apply. We also define $\mathcal{O}$ to be the set of bit-strings, $\vec{b}$, such that $O_{\vec{b}}$ is non-empty, and we define $k \coloneqq \|\mathcal{O}\|$.  $\{O_{\vec{b}} : \vec{b} \in \mathcal{O}\}$ partitions $X$ according to which combination of input constraints apply.

Given:
\begin{itemize}
     \item $c$ different input constraint \emph{proximity functions},
     $\varsigma_i : X \to [0,1]$, where $\varsigma_i$ is continuous and $\forall x \in A_i$, $\varsigma_i(x) = 0$
     
     \item $k$ different \emph{constrained predictors}, \begin{small} $G_{\vec{b}} : X \to \bigcap\limits_{\{i: b_i = 1\}} B_i$ \end{small}, one for each $\vec{b} \in \mathcal{O}$, such that the domain of each $G_{\vec{b}}$ is non-empty\footnote{For example, $G_{101}$ maps inputs from $X$ to the output region $(B_1 \cap B_3)$.  Note that $G_{0\cdots0}$ is a degenerate ``constrained'' predictor, since none of the constraints apply to it and it can map into all of $Y$.}
 \end{itemize}
we define:
\begin{itemize}
    \item $k$ different \emph{weighting functions}, $w_{\vec{b}}(x) \coloneqq \prod\limits_{\{i: b_i = 0\}} \left[\varsigma_i(x)\right] \prod\limits_{\{i: b_i = 1\}} \left[1 - \varsigma_i(x)\right]$
    
    \item a \emph{safe predictor}, $F(x) \coloneqq \frac{\sum\limits_{\vec{b} \in \mathcal{O}} w_{\vec{b}}(x) \cdot G_{\vec{b}}(x)}{\sum\limits_{\vec{b} \in \mathcal{O}} w_{\vec{b}}(x)}$
\end{itemize}

\begin{theorem*} \label{thm:thm}
 For all $i$, if $x \in A_i$, $F(x) \in B_i$.
\end{theorem*}

A formal proof of Theorem~\ref{thm:thm} is presented in Appendix~\ref{app:proof}, and can be summarized as: if an input is in $A_1$, then by construction of the proximity and weighting functions, all of the constrained predictors, $G_{\vec{b}}$, that do not map to $B_1$ will be given zero weight. Only the constrained predictors that map to $B_1$ will be given non-zero weight, and due to the convexity of $B_1$, the weighted average of the predictions will remain in $B_1$.

If all $G_{\vec{b}}$ are continuous and if there are no two input sets, $A_i$ and $A_j$, for which $(A_i \cap A_j) \subset (\partial A_i \cup \partial A_j)$ (i.e., no input constraint regions intersect only on their boundary), then $F$ is continuous. 
In the worst case, as the number of constraints grows linearly, the number of constrained predictors required to describe our safe predictor will grow exponentially.\footnote{For $c$ constraints, the maximum number of possible non-empty overlap regions is $k=2^c$.}  For real applications, however, we expect many of the constraint overlap sets, $O_{\vec{b}}$, to be empty.  Thus, any predictors that correspond to an empty set can be ignored, resulting in a much lower number of constrained predictors needed in practice. 

\begin{figure}
    \centering
    \includegraphics[width=1.0\textwidth]{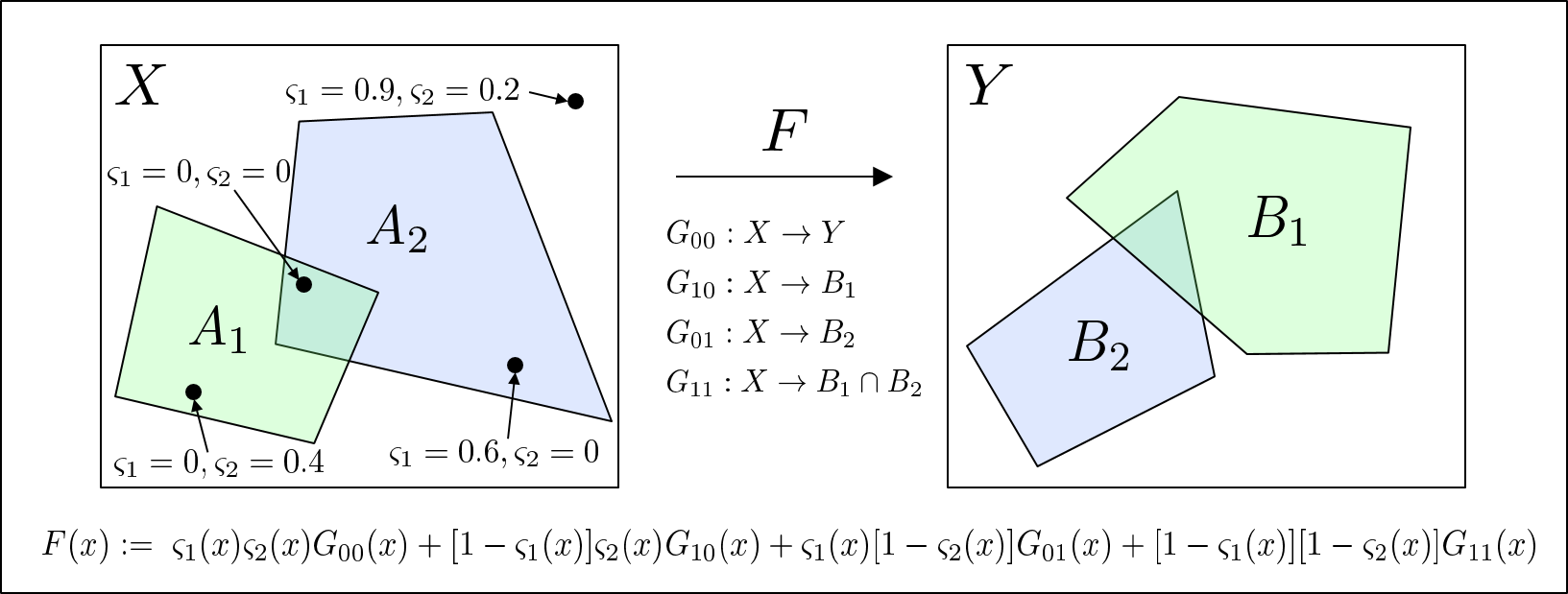}
    \caption{Notional depiction of a safe predictor with two input-output constraints.}
    \label{fig:notional_scenario}
\end{figure}

See Figure~\ref{fig:notional_scenario} for an illustrative example of how to construct $F(x)$ for a notional problem with two overlapping input-output constraints.

\subsection{Proximity Functions}
The proximity functions, $\varsigma_i$, describe how close an input, $x$, is to a given input constraint region, $A_i$, and these functions are used to compute the weights of the constrained predictors.  A desirable property for $\varsigma_i$ is for $\varsigma_i(x) \to 1$ as $d(x,A_i) \to \infty$, for some distance function,\footnote{For a set $S \subset X$, $d(x,S) \coloneqq \inf\limits_{x' \in S}d(x,x')$.} so that when an input is far from a constraint region, the constraint has little effect on the prediction for that input. A natural choice for a function that provides this property is:
\begin{small}
\begin{equation}
    \varsigma_i(x;\Sigma_i) = 1 - \exp\left[-\left(\frac{d(x,A_i)}{\sigma_1}\right)^{\sigma_2}\right], \label{eqn:proximity}
\end{equation}
\end{small}where $\Sigma_i$ is the pair of parameters $\sigma_1 \in (0,\infty)$ and $\sigma_2 \in (1,\infty)$, which could be specified using engineering judgment, or learned via optimization over training data. In our experiments in this paper, we use proximity functions of this form and learn independent parameters for each input-constrained region.  We plan to explore other choices for proximity functions in future work.

\subsection{Learning}
If we have families of differentiable functions $G_{\vec{b}}(x;\theta_{\vec{b}})$, continuously parameterized by $\theta_{\vec{b}}$, and families of $\varsigma_{i}(x;\Sigma_i)$, differentiable and continuously parameterized by $\Sigma_i$, then $F(x;\vec{\theta},\vec{\Sigma})$, where $\vec{\theta}{=}\{\theta_{\vec{b}}:\vec{b} \in \mathcal{O}\}$ and \mbox{$\vec{\Sigma}{=}\{\Sigma_i:i=1,\ldots,c \}$}, is also continuously parameterized and differentiable. We can now perform ordinary optimization techniques (e.g., gradient descent) to find the parameters of $F$ that minimize a loss function on some dataset, while still preserving the desired safety properties. Note that the safety guarantee holds no matter which parameters are chosen.  In practice, to create each $G_{\vec{b}}(x;\theta_{\vec{b}})$, we imagine choosing:
\begin{itemize}
    \item a latent space $\mathbb{R}^m$,
    \item a map \begin{small} $h_{\vec{b}} : \mathbb{R}^m \to \bigcap\limits_{\{i:b_i=1\}} B_i$ \end{small},
    \item a standard neural network architecture $g_{\vec{b}} : X \to \mathbb{R}^m$,
\end{itemize}
and then defining $G_{\vec{b}}(x;\theta_{\vec{b}}) \coloneqq h_{\vec{b}}(g_{\vec{b}}(x;\theta_{\vec{b}}))$. 

Note that this framework does not necessarily require an entirely separate network for each $\vec{b}$. In many applications, it may be useful for the constrained predictors to share earlier layers, thus learning a shared representation of the input space. Additionally, our definition of the safe predictor is general and not limited to neural networks.

In Appendix~\ref{app:synthetic}, we show an example of applying our approach to synthetic datasets in 2-D and 3-D using simple neural networks.  These examples illustrate that our safe predictor can enforce arbitrary input-output specifications with convex output constraints on neural networks and that the function we are learning is smooth.

\section{Application to Aircraft Collision Avoidance} \label{Sec:Experiment}

Aircraft collision avoidance is an application that requires strong safety guarantees. The Next-Generation Collision Avoidance System (ACAS X), which issues advisories to avoid near mid-air collisions and will have both manned (ACAS Xa) and unmanned (ACAS Xu) variants, was originally designed to select optimal advisories while minimizing disruptive alerts by solving a partially-observable Markov decision process (\cite{kochenderfer2015decision}).  The solution took the form of an extremely large look-up table mapping each possible input combination to scores for each possible advisory, and the advisory with the highest score is issued. \cite{julian2016policy} proposed compressing the policy tables using a deep neural network (DNN), which then introduced the need to verify that the DNNs met certain safety specifications.

In~\cite{jeannin2017formally}, the authors defined a desirable ``safeability'' property for ACAS X, which specified that for any given input state in the ``safeable region,'' an advisory would never be issued that would put the aircraft in a future state for which a safe advisory (i.e., an action that will prevent a collision) no longer existed. This notion is similar to the concept of control invariance (\cite{blanchini1999set}). ~\cite{julian2019verifying} created a simplified model of the ACAS Xa system (called VerticalCAS), generated DNNs to approximate the learned policy, and used Reluplex (\cite{katz2017reluplex}) to verify whether the DNNs satisfied the safeability property.  The authors found thousands of counterexamples for which their DNNs violated this property, and suggested that the construction of a network that ensures such a property remained an open problem. 

Our proposed safe predictor will ensure that any collision avoidance system will meet the safeability property by construction.  We describe in detail in Appendix~\ref{app:acas} how we apply our approach to a subset of the VerticalCAS datasets using a conservative, convex approximation of the safeability constraints. These constraints are defined such that if a given aircraft state is in the ``unsafeable region,'' $A_{\mathrm{unsafeable},i}$, for the $i^{\mathrm{th}}$ advisory, the score for that advisory must not be the highest, i.e., $x \in A_{\mathrm{unsafeable},i} \Rightarrow F^{i}(x) < \max_{j} F^{j}(x)$, where $F^{j}(x)$ is the output score for the $j^{\mathrm{th}}$ advisory. 

Results of our experiments are in Table~\ref{tab:results} where we compare a standard, unconstrained network to our safe predictor, and report the percentage accuracy (\textsc{Acc}) and violations (i.e., percentage of inputs for which the network outputs an ``unsafeable'' advisory) for each network. We train and test using PyTorch (\cite{paszke2017automatic}) with two separate datasets: one based on the previous advisory being {\it Clear of Conflict} (COC) and the other for the previous advisory {\it Climb at 1500 ft/min} (CL1500).\footnote{A description of all of the advisories is given in Table~\ref{tab:advisories} in Appendix~\ref{app:acas}.}  As expected, our safe predictor does not violate the desired safeability property. Additionally, the accuracy of our predictor is not impacted with respect to the unconstrained network, an indication that we are not losing accuracy in order to achieve safety guarantees for this example.

\begin{table*}[ht]
    \begin{small}
    \begin{sc}
    \centering
    \begin{tabular}{lccccr}
    \hline
    Network &{Acc (COC)} & {Violations (COC)} & {Acc (CL1500)} & {Violations (CL1500)}\\ \hline
    Standard & 96.87  & 0.22  & 93.89 & 0.20 \\
    Safe & 96.69  & 0.00  & 94.78 & 0.00 \\
    \end{tabular}
    \caption{Summary of percent accuracy and violations for VerticalCAS}
    \label{tab:results}
    \end{sc}
    \end{small}
    \end{table*}

\section{Discussions and Future Work}

We present an approach for designing a safe predictor that obeys a set of input-output specifications for use in safety-critical machine learning systems, and demonstrate it on a problem in aircraft collision avoidance.  The novelty of this approach is in its simplicity and guaranteed enforcement (at all stages of algorithm training) of the specifications via combinations of convex output constraints. Future work includes adapting and leveraging techniques from optimization (\cite{amos2017optnet}) and control barrier functions (\cite{cheng2019end, glotfelter2019hybrid}), and incorporating notions of adversarial robustness into the design of our safe predictor, such as extending the work of~\cite{hein2017formal, weng2018evaluating}, and~\cite{virmaux2018lipschitz} to bound the Lipschitz constant of our networks. 

\footnotesize
\bibliographystyle{plainnat}
\bibliography{references}
\normalsize

\appendix

\section{Proof of Theorem~\ref{thm:thm}} \label{app:proof}
\begin{proof}
 Fix $i$, and suppose that $x \in A_i$. Thus $\varsigma_i(x) = 0$, so for all $\vec{b} \in \mathcal{O}$ where $b_i = 0$, $w_{\vec{b}}(x) = 0$. Thus $F(x) \coloneqq \frac{\sum\limits_{ \{\vec{b} : b_i = 1\}} w_{\vec{b}}(x) \cdot G_{\vec{b}}(x)}{\sum\limits_{\{\vec{b} : b_i = 1\}} w_{\vec{b}}(x)}$.  If $b_i = 1$, $G_{\vec{b}}(x) \in B_i$, thus $F(x)$ is also in $B_i$ by the convexity of $B_i$.
\end{proof}

\section{Example on Synthetic Datasets} \label{app:synthetic}

\begin{figure}[ht!]
    \centering
    \includegraphics[width=0.77\textwidth]{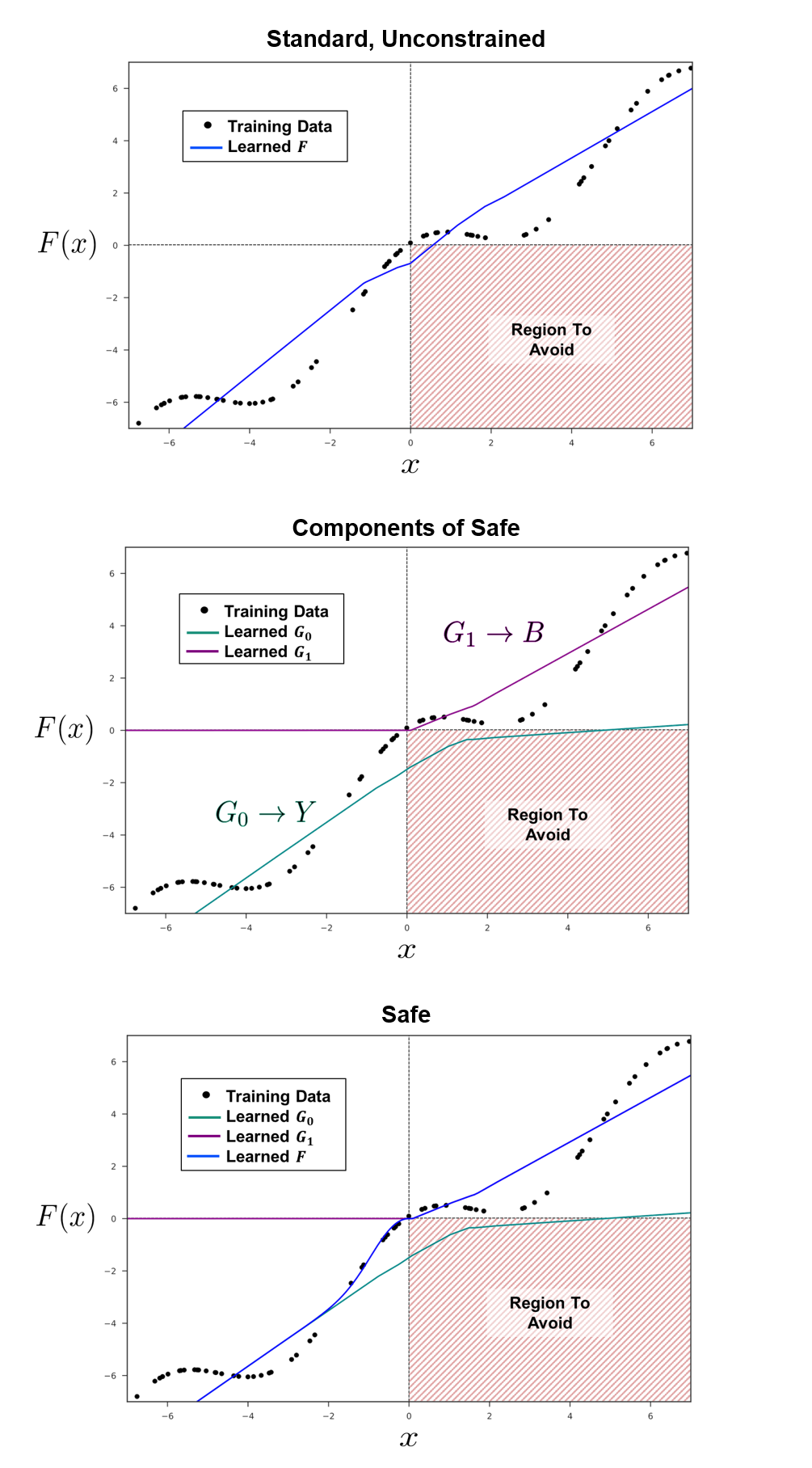}
    \caption{Example of standard, unconstrained (top) and safe, constrained (bottom) neural networks for a notional regression problem with $X \coloneqq \mathbb{R}^1, Y \coloneqq \mathbb{R}^1$, and the constraint that $x>0 \Rightarrow y=F(x)>0$.  Note that the unconstrained network (blue line in top plot) violates the constraint for the inputs that are positive and close to $0$. The middle plot shows the predictors, $G_0$ and $G_1$, that are used to construct the safe predictor, which is depicted by the blue line in the bottom plot.  Note that the safe predictor smoothly transitions from $G_0$ to $G_1$ and obeys the specified constraint, while performing similar to the unconstrained predictor in the areas outside of the constraint.}
    \label{fig:synthetic1}
\end{figure}

Figure~\ref{fig:synthetic1} depicts an example of applying our safe predictor to a notional regression problem given inputs and outputs in 1-D with one input-output constraint.  The unconstrained network has a single hidden layer of dimension 10 with rectified linear unit (ReLU) activations, followed by a fully connected layer.  For the safe predictor, the constrained predictors, $G_0$ and $G_1$, share the hidden layer but have their own fully connected layer.  Training uses a sampled subset of points from the input space, and the learned predictors are shown for the continuous input space.

\begin{figure}[ht!]
    \centering
    \includegraphics[width=1\textwidth]{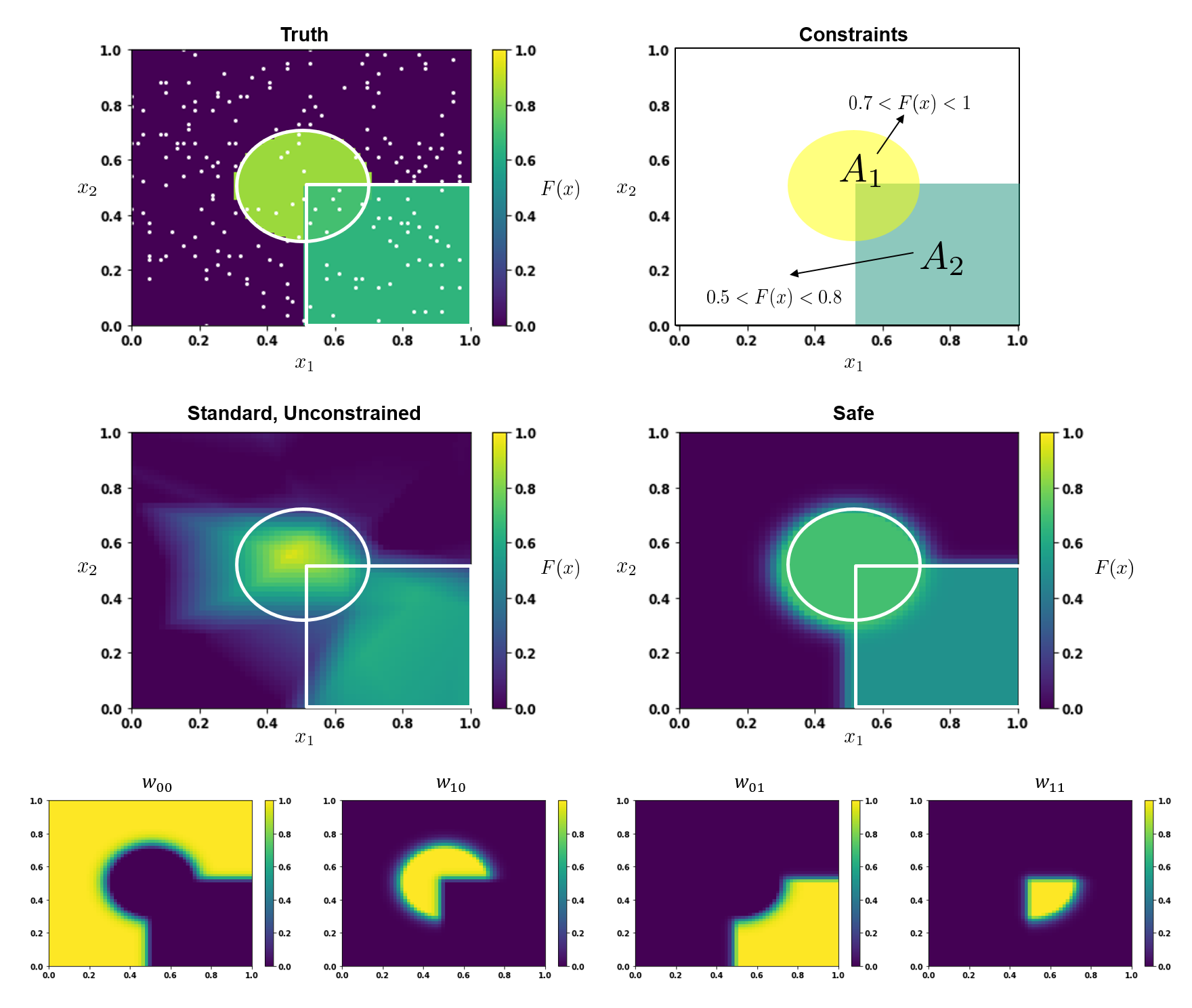}
    \caption{Example of standard, unconstrained (middle left) and safe, constrained (middle right) neural networks for a notional regression problem with $X \coloneqq  \mathbb{R}^2$, $Y \coloneqq  \mathbb{R}^1$, and the constraints that $x \in A_1 \Rightarrow 0.7<F(x)<1$, and $x \in A_2 \Rightarrow 0.5<F(x)<0.8$. Truth is shown in the top left, and the constraints are depicted in the top right.  Training samples are represented by white dots in the truth plot.  The four sets of learned weights for the constrained predictors are shown in the bottom row. Note that the unconstrained network clearly violates the constraints for some of the points in both regions. The safe predictor obeys the specified constraints, with a smooth (but quick) transition between the unconstrained and constrained regions of the input space.}
    \label{fig:synthetic2}
\end{figure}

Figure~\ref{fig:synthetic2} depicts an example of applying our safe predictor to a notional regression problem given a 2-D input and 1-D output with two overlapping constraints.  The unconstrained network has two hidden layers, each of dimension 20 and with ReLU activations, followed by a fully connected layer.  The constrained predictors, $G_{00}, G_{10}, G_{01},$ and $G_{11}$, share the hidden layers, and have their own additional hidden layer (of size 20, with ReLU) followed by a fully connected layer. Again, training uses a sampled subset of of points from the input space, and the learned predictors are shown for the continuous input space.

\section{Details of VerticalCAS Experiment} \label{app:acas}

\begin{figure}[ht!]
    \centering
    \includegraphics[width=0.6\textwidth]{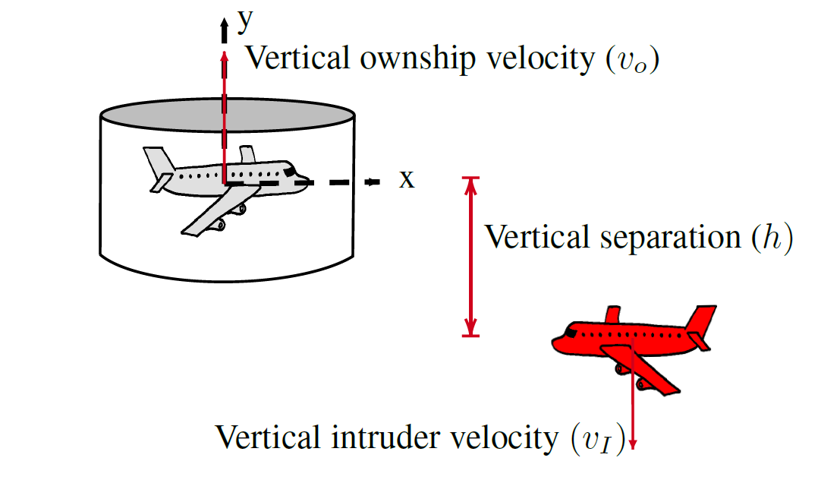}
    \caption{Depiction of three of the input variables for the VerticalCAS system, from~\cite{julian2019verifying}.}
    \label{fig:verticalcas}
\end{figure}

We start with the policies generated by the VerticalCAS system described in~\cite{julian2019verifying} and generated using their open source code.\footnote{https://github.com/sisl/VerticalCAS}  The policy tables report the scores for each of nine possible advisories (described in Table~\ref{tab:advisories}) given the current altitude, $h$, the time to loss of horizontal separation, $\tau$, the previously issued advisory, $a_{\mathrm{prev}}$, and the own aircraft and intruder aircraft vertical climb rates, $v_O$ and $v_I$, respectively.  Refer to Figure~\ref{fig:verticalcas} for a depiction of three of these variables. 

The policies are optimized using a partially observable Markov decision process to select advisories to prevent near mid air collisions (NMACs), which is defined as the intruder aircraft being within 100 vertical feet of the own aircraft at time $\tau=0$.  When in operation, the advisory with the highest score for the current input state is reported. \cite{julian2019verifying} chose to split the policy table based on $a_{\mathrm{prev}}$, and train a separate neural network for each previous advisory. Similarly, we test our approach using separate policy tables for two previous advisories: {\it Clear of Conflict} (COC) and {\it Climb at 1500 ft/min} (CL1500).

\begin{table}[ht!]
\centering
\begin{tabular}{llllll}
\hline
\textsc{Advisory} & \textsc{Description}  \\ \hline
COC & Clear of Conflict \\
DNC & Do Not Climb  \\
DND & Do Not Descent \\
DES1500 & Descent at least 1500 ft/min \\
CL1500 & Climb at least 1500 ft/min  \\
SDES1500 & Strengthen Descent to at least 1500 ft/min \\
SCL1500 & Strengthen Climb to at least 1500 ft/min \\
SDES2500 & Strengthen Descent to at least 2500 ft/min \\
SCL2500 & Strengthen Climb to at least 2500 ft/min \\ \hline
\end{tabular}
\begin{sc}
    \vspace{0.1cm}
    \caption{Description of advisories for VerticalCAS from~\cite{julian2019verifying}}
    \label{tab:advisories}
\end{sc}
\end{table}

\subsection{Safeability Constraints}

The ``safeability'' property, originally defined in~\cite{jeannin2017formally} and used to verify the safety of the VerticalCAS neural networks in~\cite{julian2019verifying}, can be encoded into a set of input-output constraints.  The ``safeable region'' for a given advisory represents the locations in the input space for which that advisory can be selected for which future advisories exist that will prevent an NMAC from occurring.  If no future advisories exist for preventing an NMAC, the advisory for the current state is considered ``unsafeable,'' and the region in the input space for which an advisory is considered unsafeable is the ``unsafeable region.''  Refer to the top plot of Figure~\ref{fig:proximity} for an example of these regions for the CL1500 advisory.  

The constraints we would thus like to enforce in our safe predictor are the following: $ x \in A_{\mathrm{unsafeable},i} \Rightarrow F^{i}(x) < \max_{j} F^{j}(x), \forall i$, where  $A_{\mathrm{unsafeable},i}$ is the unsafeable region for the $i^{\mathrm{th}}$ advisory and $F^{j}(x)$ is the output score for the $j^{\mathrm{th}}$ advisory. As is, the output regions of the safeable constraints are not convex. To convert to convex output regions, we use a conservative approximation by enforcing $F^{i}(x) = \min_{j} F^{j}(x)$, $\forall x \in A_{\mathrm{unsafeable},i}$ (i.e., ensure that the unsafeable advisory always has the lowest score).

\subsection{Proximity Functions}

\begin{figure}[h]
    \centering
    \includegraphics[width=0.75\textwidth]{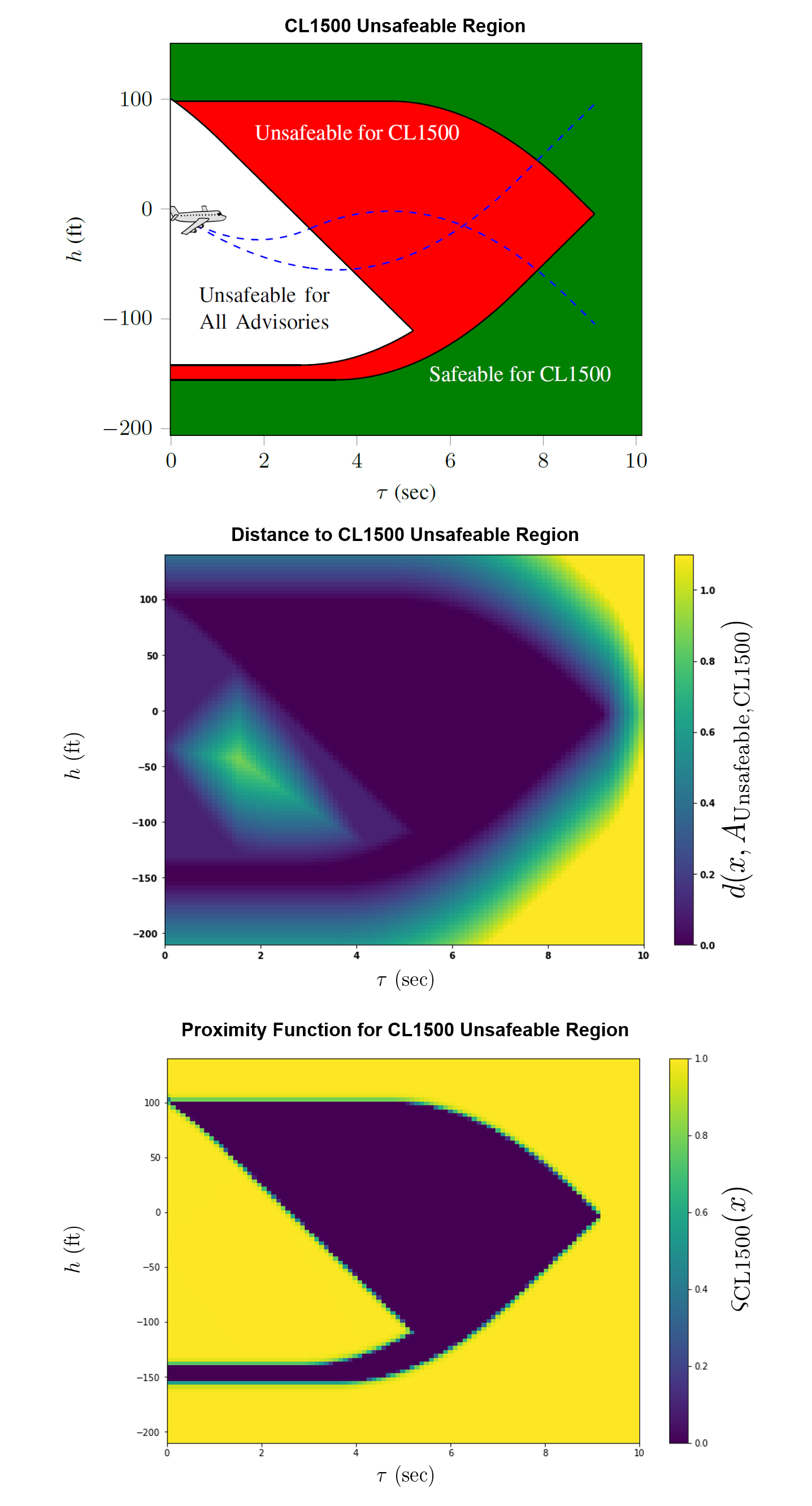}
    \caption{Top plot shows the safeable region (green) and unsafeable region (red) for the CL1500 advisory, and the region where no safeable advisories exist (white), generated in~\cite{julian2019verifying}. Middle and bottom plots depict the ``distance'' to the unsafeable region and the learned proximity function, respectively, for the CL1500 advisory, for a slice of the input space at $v_O-v_I=-180 \; \mathrm{ft/sec}$. The learned proximity function is 0 inside the unsafeable region (by construction) and quickly transitions to 1 outside of this region.}
    \label{fig:proximity}
\end{figure}

We start by generating the bounds on the unsafeable regions using open source code\footnote{https://github.com/kjulian3/Safeable} from~\cite{julian2019verifying}, then computing a ``distance function'' between points in the input space, $(v_O-v_I,h,\tau)$, and the unsafeable region for each advisory. While not true distances, these values are $0$ if and only if the data point is inside the unsafeable set, and so when they are used to produce proximity functions as in Equation \ref{eqn:proximity}, the safety properties are respected.  Examples of the unsafeable region, distance function, and proximity function (after the parameters, $\Sigma$, have been optimized during training) for the CL1500 advisory are shown in Figure~\ref{fig:proximity}.

\subsection{Structure of Predictors}
The compressed versions of the policy tables created by both~\cite{julian2016policy} for ACAS Xu and~\cite{julian2019verifying} for VerticalCAS are neural networks with six hidden-layers, 45 dimensions in each hidden layer, and ReLU activation functions. We use this same architecture for our implementation of the standard, unconstrained network. 

For our constrained predictors, we chose to use this same structure, except the first four hidden layers are shared between all of the predictors. The idea here is that we should learn a single, shared representation of the input space, while still giving each predictor some room to adapt to its own constraints. Following the shared layers, each individual constrained predictor has two additional hidden layers, and their final outputs are projected onto our convex approximation of the safe region of the output space. This is accomplished by setting the score for any unsafeable advisory, $i$, to $G_{\vec{b}}^i(x)=\min_{j} G_{\vec{b}}^j(x)-\epsilon$, where $G_{\vec{b}}^j$ is the score assigned to the $j^{\mathrm{th}}$ advisory by the constrained predictor, $G_{\vec{b}}$.  We use $\epsilon=0.0001$ in our experiments.

The number of separate predictors required to enforce the VerticalCAS safeability constraints using our approach is 30. These predictors introduce additional parameters, thus increasing the size of the network from 270 to 2880 nodes for the unconstrained and safe implementations, respectively. Our safe predictor, however, remains orders of magnitude smaller than the original look-up tables.

\subsection{Parameter Optimization}
We define our networks and perform the parameter optimization using PyTorch (\cite{paszke2017automatic}).  We optimize the parameters of both the unconstrained network and our safe predictor using the asymmetric loss function from~\cite{julian2016policy} and~\cite{julian2019verifying} to guide the network to select optimal advisories while also accurately predicting the scores from the look-up tables for all advisories.  We split each dataset using an $80/20$ train and test split, respectively, with the random seed set to $0$ using scikit-learn's (\cite{scikit-learn})  \textsc{StratifiedShuffleSplit}. The optimizer is \textsc{Adam} (\cite{kingma2014adam}) with a learning rate of $0.0003$ and batch size of $2^{16}$. The number of training epochs is set to $500$.

\end{document}